\documentclass[10pt,twocolumn,letterpaper]{article}

\usepackage{cvpr}
\usepackage{times}
\usepackage{epsfig}
\usepackage{graphicx}
\usepackage{amsmath}
\usepackage{amssymb}

\usepackage{afterpage}
\usepackage{bm}
\usepackage{booktabs}
\usepackage{fancyhdr}
\usepackage{float}
\usepackage{multirow}
\usepackage{subcaption}
\usepackage[dvipsnames,table]{xcolor}
\usepackage{array}
\newcolumntype{H}{>{\setbox0=\hbox\bgroup}c<{\egroup}@{}} 

\usepackage{siunitx}

\usepackage[pagebackref=true,breaklinks=true,letterpaper=true,colorlinks,bookmarks=false]{hyperref}

\cvprfinalcopy 

\def\cvprPaperID{****} 

\newcommand{\stufig}[5]                                       
{
        \begin{figure}[#5]
        \begin{center}
                \includegraphics[#1]{#2}
                \caption{#3}
                \label{#4}
        \end{center}
		\vspace{-.7cm}
        \end{figure}
}

\newcommand{\stufigstar}[5]                                   
{
        \begin{figure*}[#5]
        \begin{center}
                \includegraphics[#1]{#2}
                \caption{#3}
                \label{#4}
        \end{center}
		\vspace{-.7cm}
        \end{figure*}
}

\newenvironment{stusubfig}[1]
{
        \begin{figure}[#1]
        \begin{center}
}
{
        \end{center}
        \end{figure}
}

\newenvironment{stusubfig*}[1]
{
        \begin{figure*}[#1]
        \begin{center}
}
{
        \end{center}
        \end{figure*}
}

\ifcvprfinal\fi
\begin{document}

\title{On-the-Fly Adaptation of Regression Forests for Online Camera Relocalisation\vspace{-.6\baselineskip}}

\author{
\begin{tabular}{c}
\begin{tabular}{c@{\hskip 0.85cm}c@{\hskip 0.85cm}c}
Tommaso Cavallari$^1$ & Stuart Golodetz$^2$\thanks{S. Golodetz and N. Lord assert joint second authorship.} & Nicholas A. Lord$^2$$^*$ \\
Julien Valentin$^3$ & Luigi Di Stefano$^1$ & Philip H. S. Torr$^2$ \\
\end{tabular}
\\\\
\normalsize
\begin{tabular}{c}
$^1$ Department of Computer Science and Engineering, University of Bologna \\
$^2$ Department of Engineering Science, University of Oxford \\
$^3$ perceptive$^{\text{io}}$, Inc.
\end{tabular}
\\\\
\begin{tabular}{c}
$^1$ \normalsize\texttt{\{tommaso.cavallari,luigi.distefano\}@unibo.it} \\
$^2$ \normalsize\texttt{\{smg,nicklord,phst\}@robots.ox.ac.uk} \\
$^3$ \normalsize\texttt{julien@perceptiveio.com}
\vspace{-.8\baselineskip}
\end{tabular}
\end{tabular}
}

\maketitle


\begin{abstract}
Camera relocalisation is an important problem in computer vision, with applications in
simultaneous localisation and mapping, virtual/augmented reality and
navigation. Common techniques either match the current image against
keyframes with known poses coming from a tracker, or establish 2D-to-3D correspondences
between keypoints in the current image and points in the scene in order to
estimate the camera pose.
Recently, regression forests have become a popular alternative to establish
such correspondences. They achieve accurate results,
but must be trained offline on the target scene, preventing relocalisation in new
environments. In this paper, we show how to circumvent this
limitation by adapting a pre-trained forest to a new scene on the fly.
Our adapted forests achieve relocalisation performance that is on par with that of offline
forests, and our approach runs in under 150ms, making it desirable for
real-time systems that require online relocalisation.
\vspace{-1\baselineskip}
\end{abstract}

\section{Introduction}

\stufigstar{width=.8\linewidth}{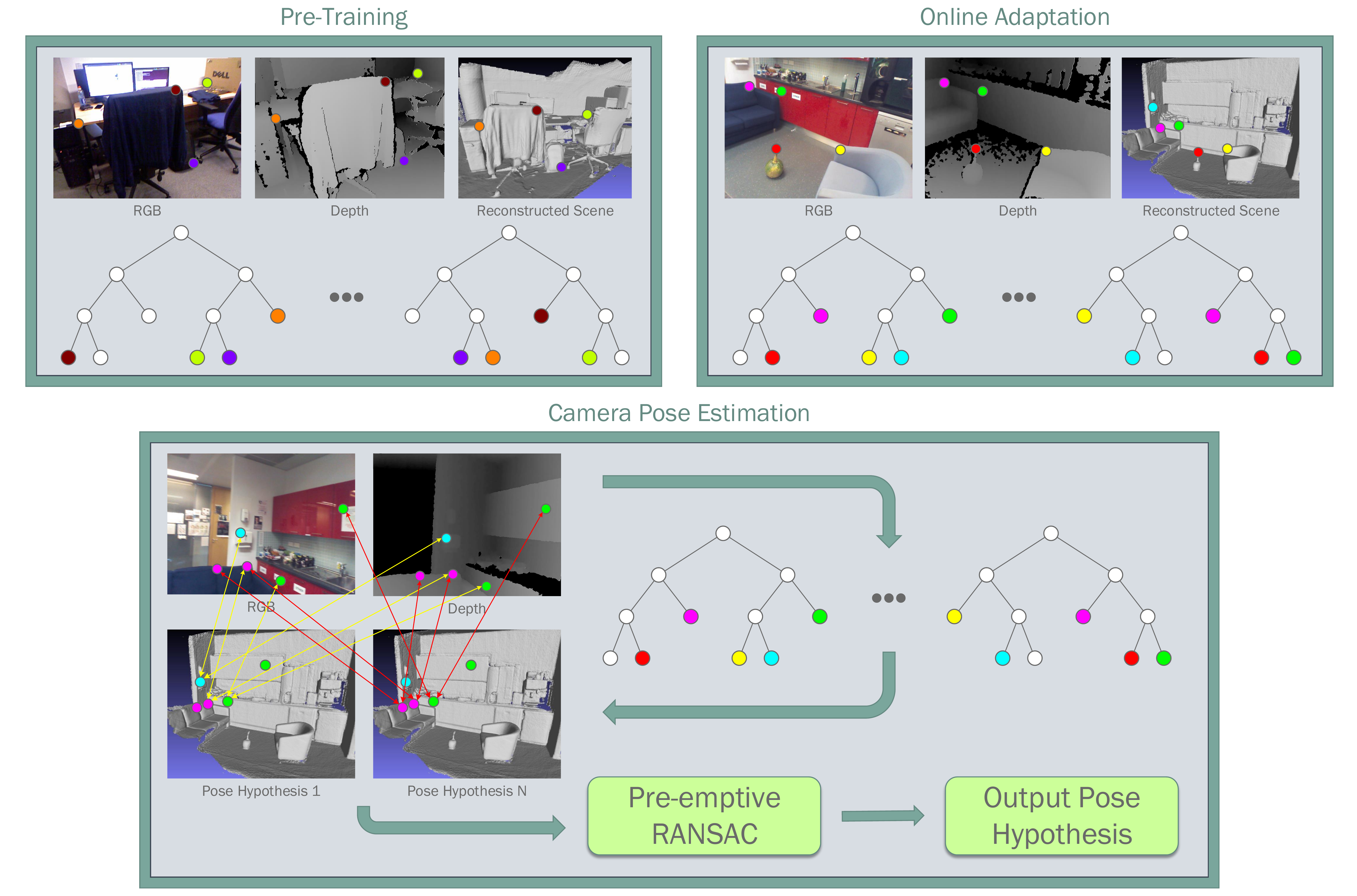}{
    \textbf{Overview of our approach}. First, we train a regression forest
    \emph{offline} to predict 2D-to-3D correspondences for a generic scene. To
    adapt this forest to a new scene, we remove the scene-specific information
    in the forest's leaves while retaining the branching structure (with learned split parameters) of the trees;
    we then refill the leaves \emph{online} using training examples from the new
    scene. The adapted forest can be deployed to predict correspondences for the
    new scene that are fed to Kabsch \cite{Kabsch1976} and RANSAC
    \cite{Fischler1981} for pose estimation.
    }{fig:pipeline}{!t}

Camera pose estimation is an important problem in computer vision, with
applications in simultaneous localisation and mapping (SLAM)
\cite{Newcombe2011,MurArtal2014,Kaehler2015}, virtual and augmented reality
\cite{Bae2016,Castle2008,Golodetz2015SPDEMO,Paucher2010,Rodas2015,Valentin2015SP} and
navigation \cite{Lee2016}. In SLAM, the camera pose is commonly initialised
upon starting reconstruction and then tracked from one frame to the next, but
tracking can easily be lost due to e.g.\ rapid movement or textureless regions
in the scene; when this happens, it is important to be able to relocalise the
camera with respect to the scene, rather than forcing the user to start the
reconstruction again from scratch. Camera relocalisation is also crucial for
loop closure when trying to build globally consistent maps
\cite{Fioraio2015,Kaehler2016,Whelan2015RSS}. Traditional approaches to camera
relocalisation have been based around one of two main paradigms:

\emph{(i) Image matching methods} match the current image from the camera
against keyframes stored in an image database (potentially with some interpolation
between keyframes where necessary). For example, Galvez-Lopez et al.\
\cite{GalvezLopez2011} describe an approach that computes a bag of binary words
based on BRIEF descriptors for the current image and compares it with bags of
binary words for keyframes in the database using an L1 score. Gee et al.\
\cite{Gee2012} estimate camera pose from a set of synthetic (i.e.\ rendered)
views of the scene. Their approach is interesting because unlike many image
matching methods, they are to some extent able to relocalise from novel
poses; however, the complexity increases linearly with the number of synthetic views
needed, which poses significant limits to practical use. Glocker et al.\
\cite{Glocker2015} encode frames using Randomised Ferns, which when evaluated on
images yield binary codes that can be matched quickly by their Hamming
distance: as noted in \cite{Li2015}, this makes their approach much faster than
\cite{Gee2012} in practice.

\emph{(ii) Keypoint-based methods} find 2D-to-3D correspondences between
keypoints in the current image and 3D scene points, so as to deploy e.g.\ a
Perspective-n-Point (PnP) algorithm \cite{Hartley2004} (on RGB data) or
the Kabsch algorithm \cite{Kabsch1976} (on RGB-D data) to generate a
number of camera pose hypotheses that can be pruned to a single hypothesis using
RANSAC \cite{Fischler1981}. For example, Williams et al.\ \cite{Williams2011}
recognise/match keypoints using an ensemble of randomised lists, and exclude
unreliable or ambiguous matches when generating hypotheses. Their approach is
fast, but needs significant memory to store the lists. Li et al.\ \cite{Li2015} use
graph matching to help distinguish between visually-similar keypoints. Their
method uses BRISK descriptors for the keypoints, and runs at around 12 FPS.
Sattler et al.\ \cite{Sattler2016} describe a large-scale localisation approach
that finds correspondences in both the 2D-to-3D and 3D-to-2D directions before
applying a 6-point DLT algorithm to compute pose hypotheses. They use a visual
vocabulary to order potential matches by how costly they will be to establish.

Some hybrid methods use both paradigms. For example, Mur-Artal et al.\
\cite{MurArtal2015} describe a relocalisation approach that initially finds pose
candidates using bag of words recognition \cite{GalvezLopez2012}, which they
incorporate into their larger ORB-SLAM system (unlike \cite{GalvezLopez2011},
they use ORB rather than BRIEF features, which they found to improve
performance). They then refine these candidate poses using PnP and RANSAC.
Valentin et al.\ \cite{Valentin2016} present an approach that finds initial pose
candidates using the combination of a retrieval forest and a multiscale
navigation graph, before refining them using continuous pose optimisation.

Several less traditional approaches have also been tried. Kendall et
al.~\cite{Kendall2015} train a convolutional neural network to directly regress
the 6D camera pose from the current image. Deng et al.~\cite{Deng2016} match a 3D
point cloud representing the scene to a local 3D point cloud constructed from a
set of query images that can be incrementally extended by the user to achieve a
successful match. Lu et al.~\cite{Lu2015} perform 3D-to-3D localisation that
reconstructs a 3D model from a short video using structure-from-motion and
matches that against the scene within a multi-task point retrieval framework.

Recently, Shotton et al.\ \cite{Shotton2013} proposed the use of a regression
forest to directly predict 3D correspondences in the scene for all pixels in the
current image. This has two key advantages over traditional keypoint-based
approaches: (i) no explicit detection, description or matching of keypoints is
required, making the approach both simpler and faster, and (ii) a significantly
larger number of points can be deployed to verify or reject camera pose
hypotheses. However, it suffers from the key limitation of needing to train a
regression forest on the scene \emph{offline} (in advance), which prevents
on-the-fly camera relocalisation.
Subsequent work has significantly improved upon the relocalisation performance
of \cite{Shotton2013}. For example, Guzman-Rivera et al.\
\cite{GuzmanRivera2014} rely on multiple regression forests to generate a number of
camera pose hypotheses, then cluster them and use the mean pose of the cluster
whose poses minimise the reconstruction error as the result. Valentin et al.\
\cite{Valentin2015RF} replace the modes used in the leaves of the forests in
\cite{Shotton2013} with mixtures of anisotropic 3D Gaussians in order to better
model uncertainties in the 3D point predictions, and show that by combining this
with continuous pose optimisation they can relocalise 40\% more frames than
\cite{Shotton2013}.
Brachmann et al.\ \cite{Brachmann2016} deploy a stacked classification-regression
forest to achieve results of a quality similar to \cite{Valentin2015RF} for
RGB-D relocalisation.
Massiceti et al.\ \cite{Massiceti2016} map between regression forests and neural
networks to try to leverage the performance benefits of neural networks for
dense regression while retaining the efficiency of random forests for
evaluation. They use robust geometric median averaging to achieve improvements
of around 7\% over \cite{Brachmann2016} for RGB localisation. However, despite
all of these advances, none of these papers remove the need to train on the
scene of interest in advance.

In this paper, we show that this need for \emph{offline} training on the scene
of interest can be overcome through \emph{online} adaptation to a new scene of a regression forest that has been pre-trained on a generic scene. We achieve genuine
on-the-fly relocalisation similar to that which can be obtained using
keyframe-based approaches \cite{Glocker2015}, but with both significantly higher
relocalisation performance in general, and the specific advantage that we can
relocalise from novel poses. Indeed, our adapted forests achieve relocalisation
performance that is competitive with offline-trained forests, whilst requiring
no pre-training on the scene of interest and relocalising in close to real time.
This makes our approach a practical and high-quality alternative to
keyframe-based methods for online relocalisation in novel scenes.

\section{Method}
\label{sec:method}

\subsection{Overview}
\label{subsec::methodoverview}

\noindent Figure~\ref{fig:pipeline} shows an overview of our approach.
Initially, we train a regression forest \emph{offline} to predict 2D-to-3D
correspondences for a \emph{generic} scene, as per \cite{Valentin2015RF}.
To adapt this forest to a new scene, we remove the
contents of the leaf nodes in the forest (i.e.\ GMM modes and associated covariance matrices)
whilst retaining the branching structure of the trees (including learned split parameters). We then adapt the forest
\emph{online} to the new scene by feeding training examples down the forest to
refill the empty leaves, dynamically learning a set of leaf distributions specific
to that scene. Thus adapted, the forest can then be used to predict
correspondences for the new scene that can be used for camera pose estimation.
Reusing the tree structures spares us from expensive offline learning on
deployment in a novel scene, allowing for relocalisation on the fly.


\subsection{Details}
\label{subsec::methoddetails}
\subsubsection{Offline Forest Training}
\label{subsubsec::forestpretraining}

Training is done as in \cite{Valentin2015RF}, greedily optimising a standard
reduction-in-spatial-variance objective over the randomised parameters of simple
threshold functions. Like \cite{Valentin2015RF}, we make use of `Depth' and
`Depth-Adaptive RGB' (`DA-RGB') features, centred at a pixel $\textbf{p}$, as
follows:
%
\begin{equation}
f^{\text{Depth}}_\Omega = D(\mathbf{p}) - D\left(\mathbf{p} + \frac{\boldsymbol{\delta}}{D(\mathbf{p})}\right)
\end{equation}
\begin{equation}
f^{\text{DA-RGB}}_\Omega = C(\mathbf{p},c) - C\left(\mathbf{p} + \frac{\boldsymbol{\delta}}{D(\mathbf{p})}, c\right)
\end{equation}
%
In this, $D(\mathbf{p})$ is the depth at $\mathbf{p}$, $C(\mathbf{p},c)$ is the
value of the $c^{\text{th}}$ colour channel at $\mathbf{p}$, and $\Omega$ is a
vector of randomly sampled feature parameters. For `Depth', the only parameter
is the 2D image-space offset $\boldsymbol{\delta}$, whereas `DA-RGB' adds the
colour channel selection parameter $c \in \{R,G,B\}$. We randomly generate 128
values of $\Omega$ for `Depth' and 128 for `DA-RGB'. We concatenate the
evaluations of these functions at each pixel of interest to yield 256D feature
vectors.

At training time, a set $S$ of training examples, each consisting of such a
feature vector $\mathbf{f} \in \mathbb{R}^{256}$, its corresponding 3D location
in the scene and its colour, is assembled via sampling from a ground truth RGB-D
video with known camera poses for each frame (obtained by tracking from depth camera
input). A random subset of these training examples is selected to train each tree in
the forest, and we then train all of the trees in parallel.

Starting from the root of each tree, we recursively partition the set of training examples in
the current node into two using a binary threshold function. To decide how to split each
node $n$, we randomly generate a set $\Theta_n$ of $512$ candidate split parameter pairs,
where each $\theta = (\phi,\tau) \in \Theta_n$ denotes the binary threshold function
\begin{equation}
\theta(\mathbf{f}) = \mathbf{f}[\phi] \ge \tau.
\end{equation}
In this, $\phi \in [0,256)$ is a randomly-chosen feature index, and $\tau \in \mathbb{R}$
is a threshold, chosen to be the value of feature $\phi$ in a randomly-chosen training example.
Examples that pass the test are routed to the right subtree of $n$; the remainder are routed
to the left.
To pick a suitable split function for $n$, we use exhaustive search to find a $\theta^{*} \in \Theta_n$
whose corresponding split function maximises the information gain that can be achieved by splitting the
training examples that reach $n$. Formally, the information gain corresponding to split parameters
$\theta \in \Theta_n$ is
\begin{equation}
V(S_n) - \displaystyle\sum_{i\in\{\text{L,R}\}} \frac{|S^i_n(\theta)|}{|S_n|} \; V(S^i_n(\theta)),
\end{equation}
in which $V(X)$ denotes the spatial variance of set $X$, and $S^L_n(\theta)$ and
$S^R_n(\theta)$ denote the left and right subsets into which the set $S_n
\subseteq S$ of training examples reaching $n$ is partitioned by the split
function denoted by $\theta$. Spatial variance is defined in terms of the
log of the determinant of the covariance of a fitted 3D Gaussian \cite{Valentin2015RF}.

For a given tree, the above process is simply recursed to a maximum depth of 15.
As in \cite{Valentin2015RF}, we train 5 trees per forest.
The (approximate, empirical) distributions in the leaves are discarded at the end of this process (we replace them during online forest adaptation, as discussed next).

\subsubsection{Online Forest Adaptation}
\label{subsubsec::forestadaptation}

\begin{stusubfig*}{!t}
	\begin{subfigure}{.47\linewidth}
		\centering
		\includegraphics[width=\linewidth]{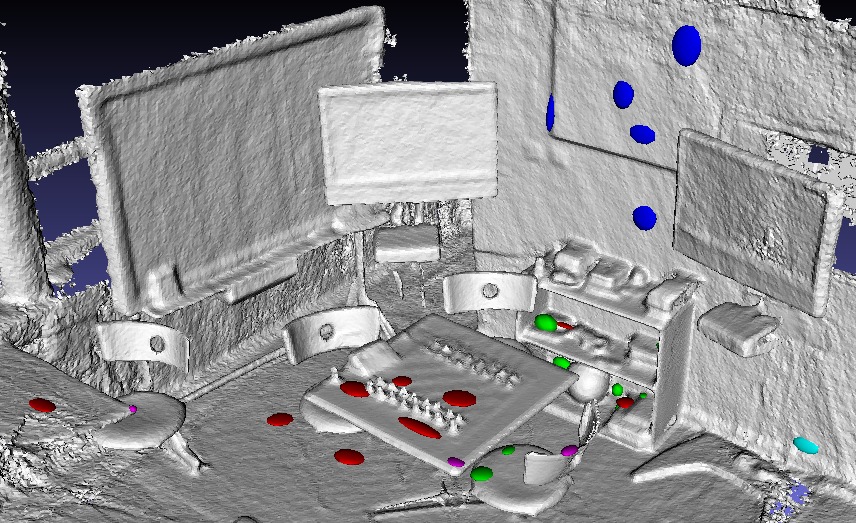}
		\caption{}
	\end{subfigure}%
	\hspace{10mm}%
	\begin{subfigure}{.47\linewidth}
		\centering
		\includegraphics[width=\linewidth]{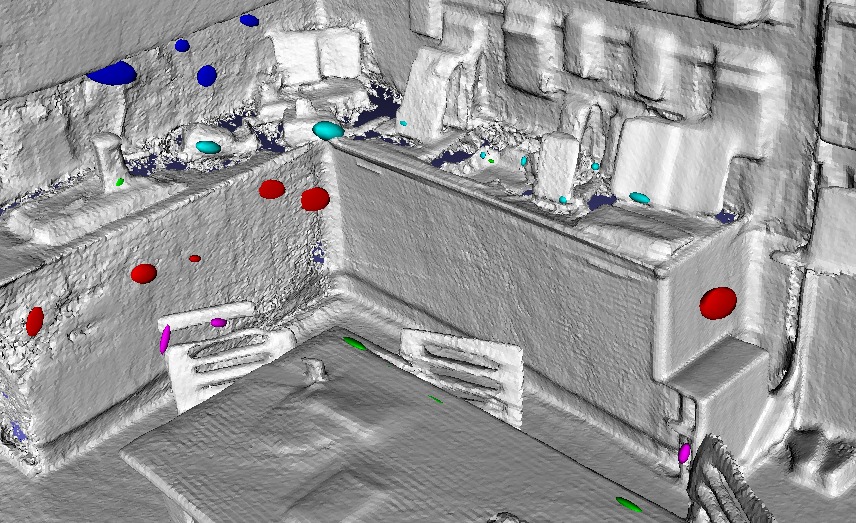}
		\caption{}
	\end{subfigure}%
\caption{An illustrative example of the effect that online adaptation has on a
    pre-trained forest: (a) shows the modal clusters present in a small number of
    randomly-selected leaves of a forest pre-trained on the \emph{Chess} scene from
    the 7-Scenes dataset \cite{Shotton2013} (the colour of each mode indicates its
    containing leaf); (b) shows the modal clusters that are added to the same leaves
    during the process of adapting the forest to the \emph{Kitchen} scene.}
\label{fig:modes}
\vspace{-1.5\baselineskip}
\end{stusubfig*}

To adapt a forest to a new environment, we replace the distributions discarded from its leaves at the end of pre-training with dynamically-updated ones drawn entirely from the new scene.
Here, we detail how the new leaf distributions used by the relocaliser are computed and updated online.

We draw inspiration from the use of reservoir sampling \cite{Vitter1985} in SemanticPaint \cite{Valentin2015SP}, which makes it possible to store an unbiased subset of an empirical distribution in a bounded amount of memory.
On initialisation, we allocate (on the GPU) a fixed-size sample reservoir for each leaf of the existing forest.
Our reservoirs contain up to $1024$ entries, each storing a 3D (world coordinate) location and an associated colour.
At runtime, we pass training examples (as per \S\ref{subsubsec::forestpretraining}) down the forest and identify the leaves to which each example is mapped.
We then add the 3D location and colour of each example to the reservoirs associated with its leaves.
To obtain the 3D locations of the training examples, we need to know the transformation that maps points from camera space to world space.
When testing on sequences from a dataset, this is trivially available as the ground truth camera pose, but in a live scenario, it will generally be obtained as the output of a fallible tracker.
To avoid corrupting the reservoirs in our forest, we avoid passing new examples down the forest when the tracking is unreliable.
We measure tracker reliability using the support vector machine (SVM) approach described in \cite{Kaehler2016}.
For frames for which a reliable camera pose \emph{is} available, we proceed as follows:
\begin{enumerate}
	\item First, we compute feature vectors for a subset of the pixels in the image, as detailed in \S\ref{subsubsec::forestpretraining}. We empirically choose our subset by subsampling densely on a regular grid with $4$-pixel spacing, i.e.\ we choose pixels $\{(4i,4j) \in [0,w) \times [0,h) : i,j \in \mathbb{N} \}$, where $w$ and $h$ are respectively the width and height of the image.
	\item Next, we pass each feature vector down the forest, adding the 3D position and colour of the corresponding scene point to the reservoir of the leaf reached in each tree. Our CUDA-based random forest implementation uses the node indexing described in~\cite{Sharp2008}.
	\item Finally, for each leaf reservoir, we cluster the contained points using a CUDA implementation of Really Quick Shift (RQS) \cite{Fulkerson2010} to find a set of modal 3D locations. We sort the clusters in each leaf in decreasing size order, and keep at most $10$ modal clusters per leaf. For each cluster we keep, we compute 3D and colour centroids, and a covariance matrix. The cluster distributions are used when estimating the likelihood of a camera pose, and also during continuous pose optimisation (see \S\ref{subsubsec:poseestimation}).
	Since running RQS over all the leaves in the forest would take too long if run in a single frame, we amortise the cost over multiple frames by updating $256$ leaves in parallel each frame in round-robin fashion. A typical forest contains around \num[group-separator={,}]{42000} leaves, so each leaf is updated roughly once every $6$s.
\end{enumerate}
The aforementioned reservoir size, number of modal clusters per leaf and number of leaves to update per frame were determined empirically to achieve online processing rates.

Figure~\ref{fig:modes} illustrates the effect that
online adaptation has on a pre-trained forest: (a) shows the modal clusters
present in a few randomly-selected leaves of a forest pre-trained on
the \emph{Chess} scene from the 7-Scenes dataset \cite{Shotton2013}; (b) shows
the modal clusters that are added to the same leaves during the process of
adapting the forest to the \emph{Kitchen} scene. Note that whilst the positions
of the predicted modes have (unsurprisingly) completely changed, the split
functions in the forest's branch nodes (which we preserve) still do a good
job of routing similar parts of the scene into the same leaves, enabling
effective sampling of 2D-to-3D correspondences for camera pose estimation.

\subsubsection{Camera Pose Estimation}
\label{subsubsec:poseestimation}

As in \cite{Valentin2015RF}, camera pose estimation is based on the preemptive, locally-optimised RANSAC of \cite{Chum2003}.
We begin by randomly generating an initial set of up to $1024$ pose hypotheses.
A pose hypothesis $H \in \mathbf{SE}(3)$ is a transform that maps points in camera space to world space.
To generate each pose hypothesis, we apply the Kabsch algorithm \cite{Kabsch1976} to $3$ point pairs of the form $(\mathbf{x}_i^\mathcal{C}, \mathbf{x}_i^\mathcal{W})$, where $\mathbf{x}_i^\mathcal{C} = D(\mathbf{u}_i) K^{-1} (\mathbf{u}_i^\top,1)$ is obtained by back-projecting a randomly-chosen point $\mathbf{u}_i$ in the live depth image $D$ into camera space, and $\mathbf{x}_i^\mathcal{W}$ is a corresponding scene point in world space, randomly sampled from $M(\mathbf{u}_i)$, the modes of the leaves to which the forest maps $\mathbf{u}_i$.
In this, $K$ is the intrinsic calibration matrix for the depth camera.
Before accepting a hypothesis, we subject it to a series of checks:
\begin{enumerate}
	\item First, we randomly choose one of the three point pairs $(\mathbf{x}_i^\mathcal{C},\mathbf{x}_i^\mathcal{W})$ and compare the RGB colour of the corresponding pixel $\mathbf{u}_i$ in the colour input image to the colour centroid of the mode (see \S\ref{subsubsec::forestadaptation}) from which we sampled $\mathbf{x}_i^\mathcal{W}$. We reject the hypothesis iff the L0 distance between the two exceeds a threshold.
	\item Next, we check that the three hypothesised scene points are sufficiently far from each other. We reject the hypothesis iff the minimum distance between any pair of points is less than $30$cm.
	\item Finally, we check that the distances between all scene point pairs and their corresponding back-projected depth point pairs are sufficiently similar, i.e.\ that the hypothesised transform is `rigid enough'. We reject the hypothesis iff this is not the case.
\end{enumerate}
If a hypothesis gets rejected by one of the checks, we try to generate an alternative hypothesis to replace it.
In practice, we use $1024$ dedicated threads, each of which attempts to generate a single hypothesis.
Each thread continues generating hypotheses until either (a) it finds a hypothesis that passes all of the checks, or (b) a maximum number of iterations is reached.
We proceed with however many hypotheses we obtain by the end of this process.

Having generated our large initial set of hypotheses, we next aggressively cut it down to a much smaller size by scoring each hypothesis and keeping the $64$ lowest-energy transforms (if there are fewer than $64$ hypotheses, we keep all of them).
To score the hypotheses, we first select an initial set $I = \{i\}$ of $500$ pixel indices in $D$, and back-project the denoted pixels $\mathbf{u}_i$ to corresponding points $\mathbf{x}_i^\mathcal{C}$ in camera space as described above.
We then score each hypothesis $H$ by summing the Mahalanobis distances between the transformations of each $\mathbf{x}_i^\mathcal{C}$ under $H$ and their nearest modes:
\begin{equation}
E(H) = \sum_{i \in I} \left( \min_{(\boldsymbol{\mu},\Sigma) \in M(\mathbf{u}_i)} \left\| \Sigma^{-\frac{1}{2}} (H\mathbf{x}_i^\mathcal{C} - \boldsymbol{\mu}) \right\| \right)
\end{equation}
After this initial cull, we use pre-emptive RANSAC to prune the remaining $\le 64$ hypotheses to a single, final hypothesis.
We iteratively (i) expand the sample set $I$ (by adding $500$ new pixels each time), (ii) refine the pose candidates via Levenberg-Marquardt optimisation \cite{Levenberg1944,Marquardt1963} of the energy function $E$, (iii) re-evaluate and re-score the hypotheses, and (iv) discard the worse half.
In practice, the actual optimisation is performed not in $\mathbf{SE}(3)$, where it would be hard to do, but in the corresponding Lie algebra, $\mathfrak{se}(3)$.
The details of this process can be found in \cite{Valentin2015RF}, and a longer explanation of Lie algebras can be found in \cite{Strasdat2012}.

This process yields a single pose hypothesis, which we can then return if desired.
In practice, however, further pose refinement is sometimes possible.
For example, if our relocaliser is integrated into an open-source 3D reconstruction framework such as InfiniTAM \cite{Kaehler2016}, we can attempt to refine the pose further using ICP \cite{Besl1992}.
Since tasks such as 3D reconstruction are one of the key applications of our approach, we report results both with and without ICP in Table~\ref{tbl:adaptationperformance}.


\section{Experiments}

We perform both quantitative and qualitative experiments to evaluate our
approach. In \S\ref{subsec:adaptationperformance}, we compare our
\emph{adaptive} approach to state-of-the-art \emph{offline} relocalisers
that have been trained directly on the scene of interest. We show that our adapted
forests achieve competitive relocalisation performance despite being trained
on very different scenes, enabling their use for \emph{online}
relocalisation. In \S\ref{subsec:trackinglossrecovery}, we show that we can perform this adaptation on-the-fly from live sequences, allowing us to
support tracking loss recovery in interactive scenarios.
In \S\ref{subsec:novelposes}, we evaluate how well our approach
generalises to novel poses in comparison to a keyframe-based random fern
relocaliser based on \cite{Glocker2015}. This relocaliser is also practical for
on-the-fly relocalisation (hence its use in InfiniTAM \cite{Kaehler2016}), but
its use of keyframes prevents it from generalising well to novel poses.
By contrast, we are able to relocalise well even from poses that are quite far
away from the training trajectory. Finally, in \S\ref{subsec:timings}, we
compare the speed of our approach with random ferns during both normal operation
(i.e.\ when the scene is being successfully tracked) and relocalisation.
Our approach is slower than random ferns, but remains close to real-time
and achieves much higher relocalisation performance.
Further analysis can be found in the supplementary material.

\subsection{Adaptation Performance}
\label{subsec:adaptationperformance}

\begin{table*}[!t]
    \centering
    \scriptsize
    \rowcolors{2}{white}{gray!25}
    \begin{tabular}{cccccccccc}
        \toprule
        ~ & & \multicolumn{7}{c}{\textbf{Relocalisation Performance on Test Scene}} \\
        \rowcolor{white}
        \multirow{-2}{*}{\textbf{Training Scene}} & & \textbf{Chess} & \textbf{Fire} & \textbf{Heads} & \textbf{Office} & \textbf{Pumpkin} & \textbf{Kitchen} & \textbf{Stairs} & \textbf{Average (all scenes)} \\
        \midrule
        \cellcolor{white}                                       & Reloc & 99.8\% & 95.7\% & 95.5\% & 91.7\% & 82.8\% & 77.9\% & 25.8\% & 81.3\% \\
        \cellcolor{white}\multirow{-2}{*}{Chess}                & + ICP & 99.9\% & 97.8\% & 99.5\% & 94.1\% & 91.3\% & 83.3\% & 28.4\% & 84.9\% \\
        \cellcolor{white}                                       & Reloc & 98.4\% & 96.9\% & 98.2\% & 89.7\% & 80.5\% & 71.9\% & 28.6\% & 80.6\% \\
        \cellcolor{white}\multirow{-2}{*}{Fire}                 & + ICP & 99.1\% & 99.2\% & 99.9\% & 92.1\% & 89.1\% & 81.7\% & 31.0\% & 84.6\% \\
        \cellcolor{white}                                       & Reloc & 98.0\% & 91.7\% & 100\%  & 73.1\% & 77.5\% & 67.1\% & 21.8\% & 75.6\% \\
        \cellcolor{white}\multirow{-2}{*}{Heads}                & + ICP & 99.3\% & 92.3\% & 100\%  & 81.1\% & 87.7\% & 82.0\% & 31.9\% & 82.0\% \\
        \cellcolor{white}                                       & Reloc & 99.2\% & 96.5\% & 99.7\% & 97.6\% & 84.0\% & 81.7\% & 33.6\% & 84.6\% \\
        \cellcolor{white}\multirow{-2}{*}{Office}               & + ICP & 99.4\% & 99.0\% & 100\%  & 98.2\% & 91.2\% & 87.0\% & 35.0\% & 87.1\% \\
        \cellcolor{white}                                       & Reloc & 97.5\% & 94.9\% & 96.9\% & 82.7\% & 83.5\% & 70.4\% & 30.7\% & 75.5\% \\
        \cellcolor{white}\multirow{-2}{*}{Pumpkin}              & + ICP & 98.9\% & 97.6\% & 99.4\% & 86.9\% & 91.2\% & 82.3\% & 32.4\% & 84.1\% \\
        \cellcolor{white}                                       & Reloc & 99.9\% & 95.4\% & 98.0\% & 93.3\% & 83.2\% & 86.0\% & 28.2\% & 83.4\% \\
        \cellcolor{white}\multirow{-2}{*}{Kitchen}              & + ICP & 99.9\% & 98.2\% & 100\%  & 94.5\% & 90.4\% & 88.1\% & 31.3\% & 86.1\% \\
        \cellcolor{white}                                       & Reloc & 97.3\% & 95.4\% & 97.9\% & 90.8\% & 80.6\% & 74.5\% & 45.7\% & 83.2\% \\
        \cellcolor{white}\multirow{-2}{*}{Stairs}               & + ICP & 98.0\% & 97.4\% & 99.8\% & 92.1\% & 89.5\% & 81.0\% & 46.6\% & 86.3\% \\
        \cellcolor{white}                                       & Reloc & 97.3\% & 95.7\% & 97.3\% & 83.7\% & 85.3\% & 71.8\% & 24.3\% & 79.3\% \\
        \cellcolor{white}\multirow{-2}{*}{Ours (Author's Desk)} & + ICP & 99.2\% & 97.7\% & 100\%  & 88.2\% & 90.6\% & 82.6\% & 31.0\% & 84.2\% \\
        \midrule
        \cellcolor{white}                                       & Reloc & 98.4\% & 95.3\% & 97.9\% & 87.8\% & 82.2\% & 75.2\% & 29.8\% & 80.9\% \\
        \cellcolor{white}\multirow{-2}{*}{Average}              & + ICP & 99.2\% & 97.4\% & 99.8\% & 90.9\% & 90.1\% & 83.5\% & 33.5\% & 84.9\% \\
        \bottomrule
    \end{tabular}
    \vspace{1mm}
    \caption{
        The performance of our \emph{adaptive} approach after pre-training on various scenes of the 7-Scenes dataset \cite{Shotton2013}.
        We show the scene used to pre-train the forest in each version of our approach in the left column. The pre-trained forests are adapted \emph{online} for the test scene, as described in the main text.
        The percentages denote proportions of test frames with $\le 5$cm translational error and $\le 5^\circ$ angular error.
        }
    \label{tbl:adaptationperformance}
	\vspace{-\baselineskip}
\end{table*}

In evaluating the extent to which we are able to adapt a regression forest that
has been pre-trained on a different scene to the scene of interest, we seek to
answer two questions. First, how does an adapted forest compare to one that has
been pre-trained offline on the target scene? Second, to what extent does an
adapted forest's performance depend on the scene on which it has been
pre-trained? To answer both of these questions, we compare the performances of
adapted forests pre-trained on a variety of scenes (each scene from the 7-Scenes
dataset \cite{Shotton2013}, plus a novel scene containing the first author's
desk) to the performances of forests trained offline on the scene of interest
using state-of-the-art approaches
\cite{Shotton2013,GuzmanRivera2014,Valentin2015RF,Brachmann2016}.

The exact testing procedure we use for our approach is as follows. First, we
pre-train a forest on a generic scene and remove the contents of its leaves, as
described in \S\ref{sec:method}: this process runs \emph{offline} over a number
of hours or even days (but we only need to do it once). Next, we adapt the
forest by feeding it new examples from a training sequence captured on the scene
of interest: this runs \emph{online} at frame rates (in a real system, this allows
us to start relocalising almost immediately whilst training carries on in the
background, as we show in \S\ref{subsec:trackinglossrecovery}). Finally, we test
the adapted forest by using it to relocalise from every frame of a separate testing
sequence captured on the scene of interest.

As shown in Table~\ref{tbl:adaptationperformance}, the results are very accurate.
Whilst there are certainly some variations in the performance achieved by
adapted forests pre-trained on different scenes (in particular, forests trained
on the \emph{Heads} and \emph{Pumpkin} scenes from the dataset are slightly
worse), the differences are not profound: in particular, relocalisation
performance seems to be more tightly coupled to the difficulty of the scene of
interest than to the scene on which the forest was pre-trained. Notably, all of
our adapted forests achieve results that are within striking distance of the
state-of-the-art \emph{offline} methods (Table~\ref{tbl:comparativeperformance}), and
are considerably better than those that can be achieved by online competitors
such as the keyframe-based random fern relocaliser implemented in InfiniTAM
\cite{Glocker2015,Kaehler2016} (see \S\ref{subsec:novelposes}). Nevertheless,
there is clearly a trade-off to be made here between performance and
practicality: pre-training on the scene of interest is impractical for
on-the-fly relocalisation, but achieves somewhat better results, probably due to
the opportunity afforded to adapt the structure of the forest to the target
scene.

This drop in performance in exchange for practicality can be mitigated to some
extent by refining our relocaliser's pose estimates using the ICP-based tracker
\cite{Besl1992} in InfiniTAM \cite{Kaehler2015}. Valentin et al.\
\cite{Valentin2015RF} observe that the 5cm/5$^\circ$ error metric commonly used
to evaluate relocalisers is `fairly strict and should allow any robust
model-based tracker to resume'. In practice, ICP-based tracking is in many cases
able to resume from initial poses with even greater error: indeed, as
Table~\ref{tbl:adaptationperformance} shows, with ICP refinement enabled, we are
able to relocalise from a significantly higher proportion of test frames. Whilst
ICP could clearly also be used to refine the results of offline methods, what is
important in this case is that ICP is fast and does not add significantly to the
overall runtime of our approach, which remains close to real time. As such,
refining \emph{our} pose estimates using ICP yields a high-quality relocaliser
that is still practical for online use.

\iftrue
\begin{table}[!t]
\centering
\scriptsize
\begin{tabular}{cHcccccc}
\toprule
\textbf{Scene} & \textbf{Testing frames} & \cite{Shotton2013} & \cite{GuzmanRivera2014} & \cite{Valentin2015RF} & \cite{Brachmann2016} & \textbf{Us} & \textbf{Us+ICP} \\
\midrule
Chess   & 2000 & 92.6\% &  96\%  & 99.4\% & 99.6\% & 99.2\% & 99.4\% \\
Fire    & 2000 & 82.9\% &  90\%  & 94.6\% & 94.0\% & 96.5\% & 99.0\% \\
Heads   & 1000 & 49.4\% &  56\%  & 95.9\% & 89.3\% & 99.7\% &  100\% \\
Office  & 4000 & 74.9\% &  92\%  & 97.0\% & 93.4\% & 97.6\% & 98.2\% \\
Pumpkin & 2000 & 73.7\% &  80\%  & 85.1\% & 77.6\% & 84.0\% & 91.2\% \\
Kitchen & 5000 & 71.8\% &  86\%  & 89.3\% & 91.1\% & 81.7\% & 87.0\% \\
Stairs  & 1000 & 27.8\% &  55\%  & 63.4\% & 71.7\% & 33.6\% & 35.0\% \\
\midrule
Average & --   & 67.6\% & 79.3\% & 89.5\% & 88.1\% & 84.6\% & 87.1\% \\
\bottomrule
\end{tabular}
\vspace{1mm}
\caption{Comparing our \emph{adaptive} approach to state-of-the-art
    \emph{offline} methods on the 7-Scenes dataset \cite{Shotton2013} (the percentages denote
    proportions of test frames with $\le 5$cm translation error and $\le
    5^{\circ}$ angular error). For our method, we report the results obtained by
    adapting a forest pre-trained on the \emph{Office} sequence (from
    Table~\ref{tbl:adaptationperformance}). We are competitive with, and
    sometimes better than, the offline methods, without needing to pre-train
    on the test scene.}
\label{tbl:comparativeperformance}
\vspace{-\baselineskip}
\end{table}
\fi

\subsection{Tracking Loss Recovery}
\label{subsec:trackinglossrecovery}

In \S\ref{subsec:adaptationperformance}, we investigated our ability to adapt a forest to a new scene by filling its leaves with data from a training sequence for that scene, before testing the adapted forest on a separate testing sequence shot on the same scene.
Here, we quantify our ability to perform this adaptation \emph{on the fly} by filling the leaves frame-by-frame from the testing sequence: this allows recovery from tracking loss in an interactive scenario without the need for prior training on anything other than the live sequence, making our approach extremely convenient for tasks such as interactive 3D reconstruction.

Our testing procedure is as follows: at each new frame (except the first), we assume that tracking has failed, and try to relocalise using the forest we have available at that point; we record whether or not this succeeds. Regardless, we then restore the ground truth camera pose (or the tracked camera pose, in a live sequence) and, provided tracking hasn't actually failed, use examples from the current frame to continue training the forest. As Figure~\ref{fig:trackinglossrecovery} shows, we are able to start relocalising almost immediately in a live sequence (in a matter of frames, typically 4--6 are enough). Subsequent performance then varies based on the difficulty of the sequence, but rarely drops below $80\%$, except for the challenging \emph{Stairs} sequence. This makes our approach highly practical for interactive relocalisation, something we also show in our supplementary video.

\stufig{width=\linewidth}{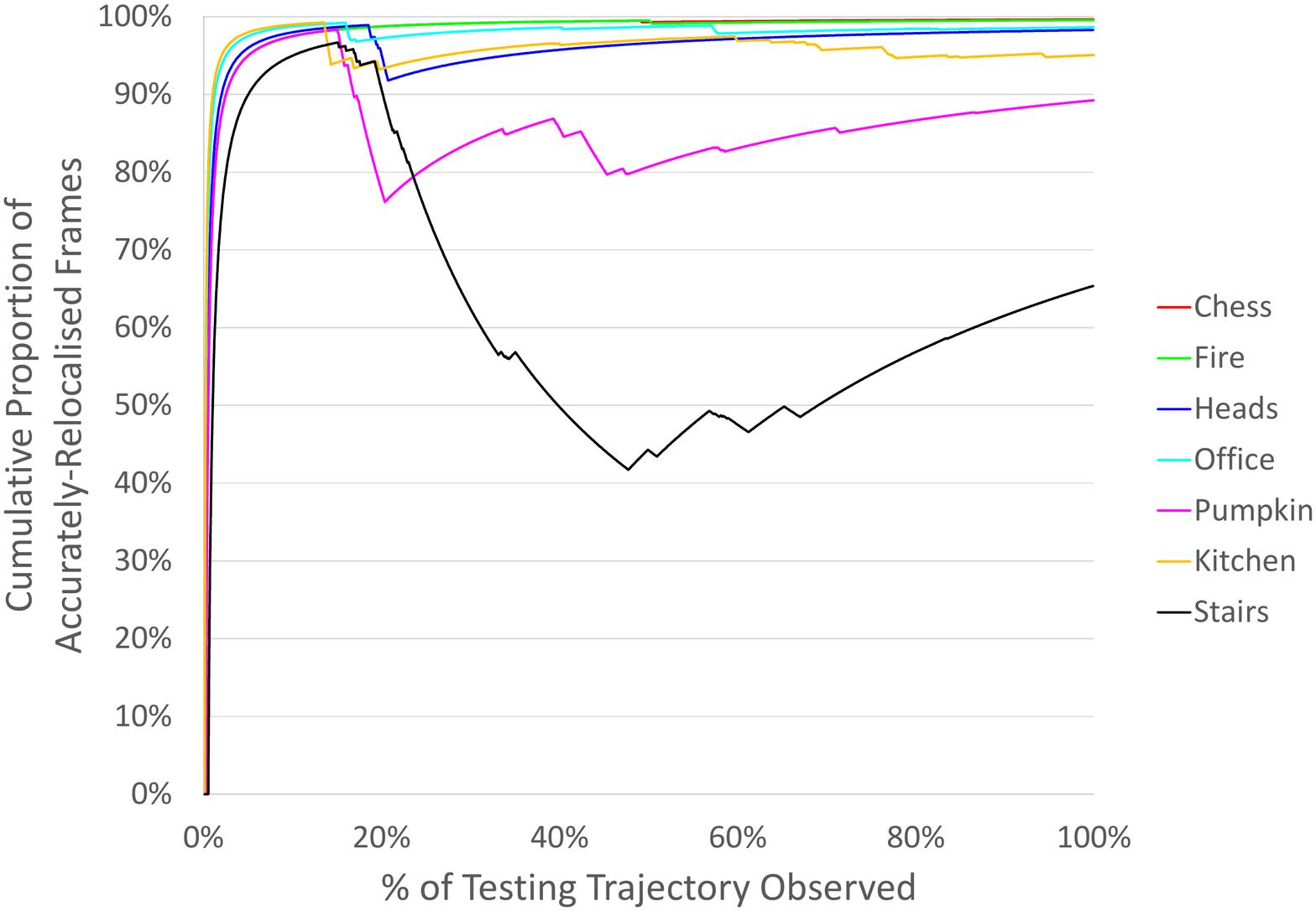}{The performance of our approach for tracking loss recovery (\S\ref{subsec:trackinglossrecovery}). Filling the leaves of a forest pre-trained on \emph{Office} frame-by-frame \emph{directly} from the \emph{testing} sequence, we are able to start relocalising almost immediately in new scenes. This makes our approach highly practical in interactive scenarios such as 3D reconstruction.}{fig:trackinglossrecovery}{!t}

\subsection{Generalisation to Novel Poses}
\label{subsec:novelposes}

\stufig{width=\linewidth}{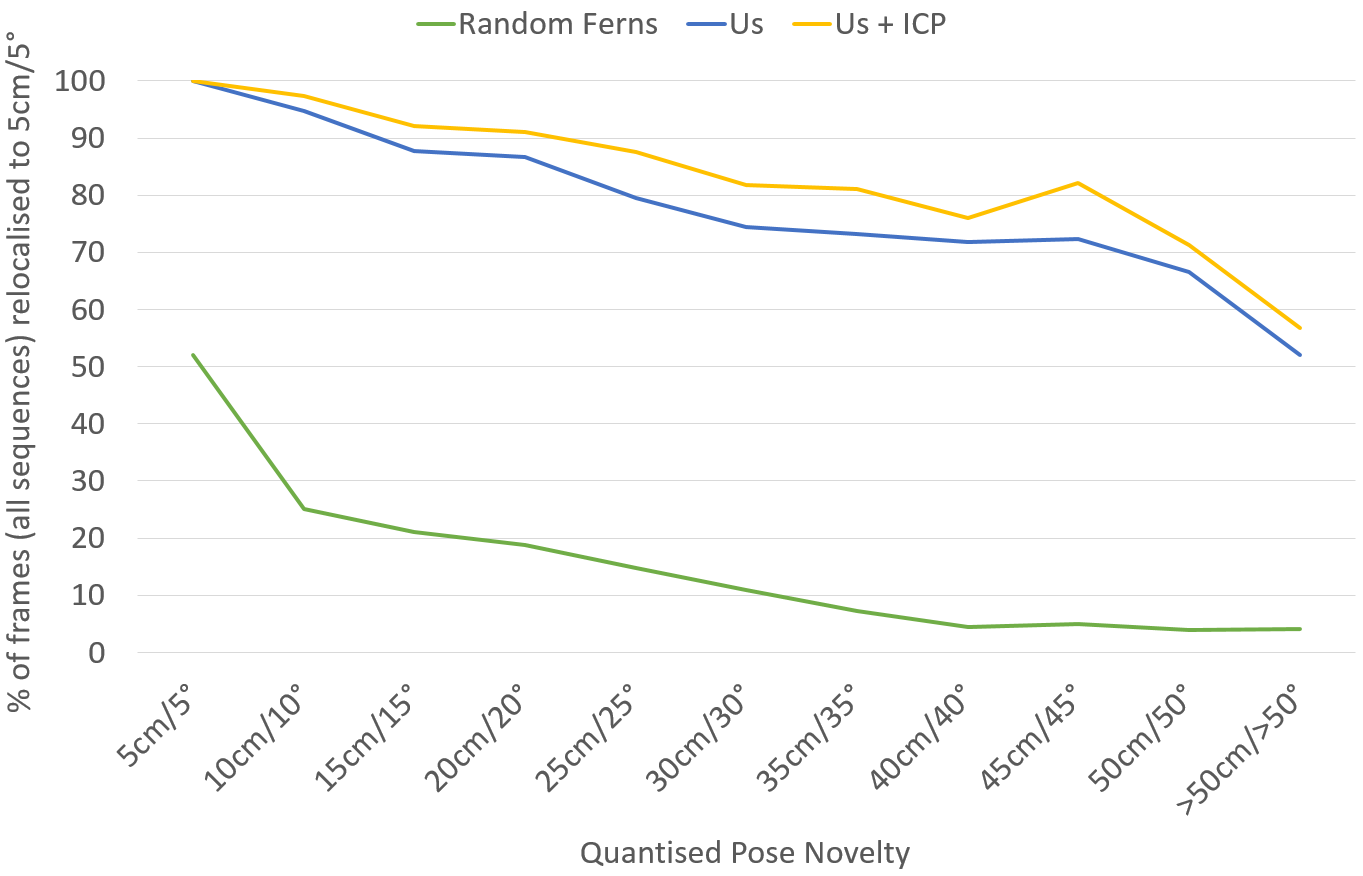}{Evaluating how well our approach generalises to novel poses in comparison to a keyframe-based random fern relocaliser based on \cite{Glocker2015}. The performance decay experienced as test poses get further from the training trajectory is much less severe with our approach than with random ferns.}{fig:novelposes-graph}{!t}

\begin{figure}[!t]
\vspace{\baselineskip}
\includegraphics[width=\linewidth]{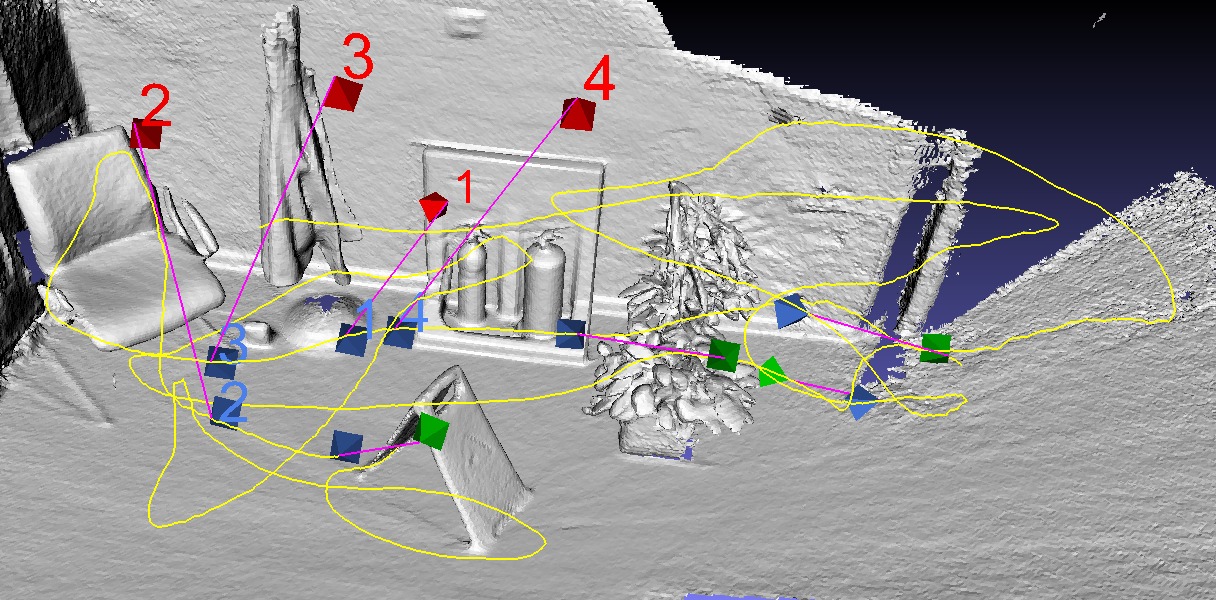}
\includegraphics[width=\linewidth]{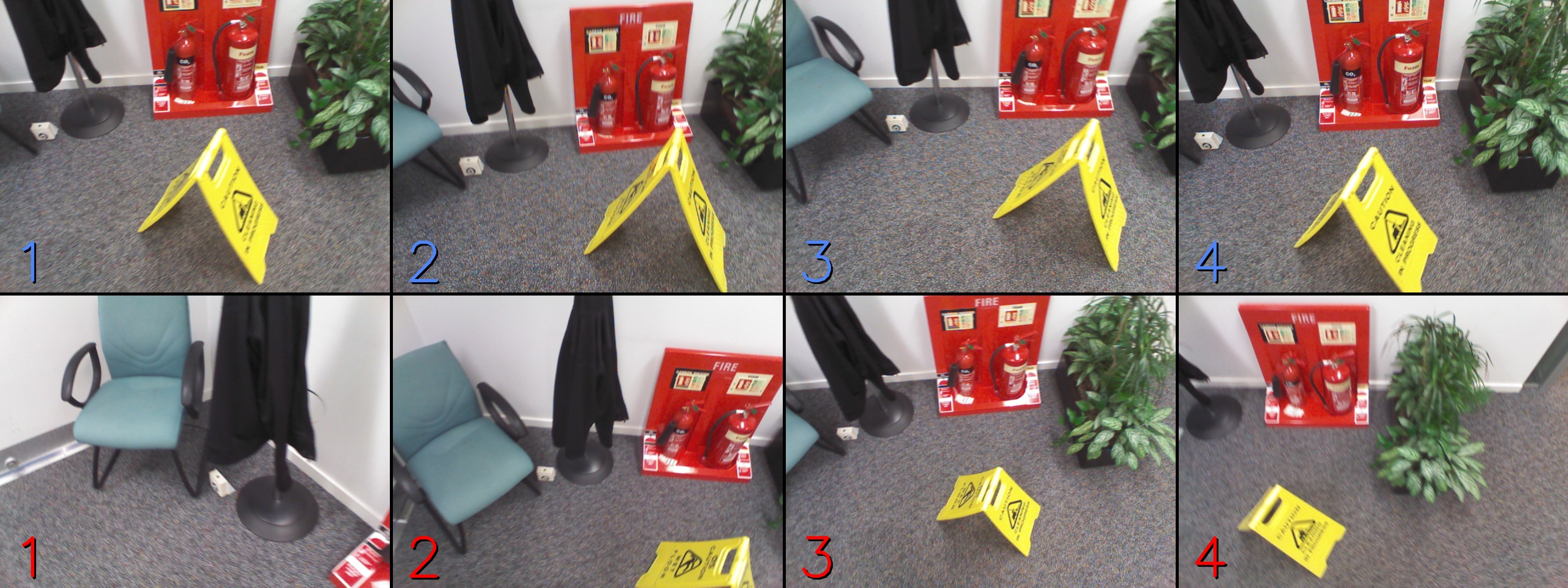}
\caption{A qualitative example of novel poses from which we are able to
    relocalise to within 5cm/5$^\circ$ on the \emph{Fire} sequence from 7-Scenes
    \cite{Shotton2013}. Pose novelty measures the distance of a test pose from a
    nearby pose (blue) on the training trajectory (yellow). We can relocalise from
    both easy poses (up to 35cm/35$^\circ$ from the training trajectory, green) and
    hard poses ($>$ 35cm/35$^\circ$, red). The images below the main figure show
    views of the scene from the training poses and testing poses indicated.}
\label{fig:novelposes-example}
\vspace{-\baselineskip}
\end{figure}

To evaluate how well our approach generalises to novel poses, we examine how the
proportion of frames we can relocalise decreases as the distance of the (ground
truth) test poses from the training trajectory increases. We compare our
approach with the keyframe-based relocaliser in InfiniTAM \cite{Kaehler2016},
which is based on the random fern approach of Glocker et al.\
\cite{Glocker2015}. Relocalisation from novel poses is a well-known failure case
of keyframe-based methods, so we would expect the random fern approach to
perform poorly away from the training trajectory; by contrast, it is interesting
to see the extent to which our approach can relocalise from a wide range of
novel poses.

We perform the comparison separately for each 7-Scenes sequence, and then
aggregate the results. For each sequence, we first group the test poses into
bins by pose novelty. Each bin is specified in terms of a maximum translation
and rotation difference of a test pose with respect to the training trajectory
(for example, poses that are within 5cm and 5$^\circ$ of any training pose are
assigned to the first bin, remaining poses that are within 10cm and 10$^\circ$
are assigned to the second bin, etc.). We then determine the proportion of the
test poses in each bin for which it is possible to relocalise to within $5$cm
translational error and $5^\circ$ angular error using (a) the random fern
approach, (b) our approach without ICP and (c) our approach with ICP. As shown
in Figure~\ref{fig:novelposes-graph}, the decay in performance experienced as
the test poses get further from the training trajectory is much less severe with
our approach than with random ferns.

A qualitative example of our ability to relocalise from novel poses is shown in
Figure~\ref{fig:novelposes-example}. In the main figure, we show a range of test
poses from which we can relocalise in the \emph{Fire} scene, linking them to
nearby poses on the training trajectory so as to illustrate their novelty in
comparison to poses on which we have trained. The most difficult of these test
poses are also shown in the images below alongside their nearby training poses,
visually illustrating the significant differences between the two.

As Figures~\ref{fig:novelposes-graph} and \ref{fig:novelposes-example}
illustrate, we are already quite effective at relocalising from poses
that are significantly different from those on which we have trained;
nevertheless, further improvements seem possible. For example, one interesting
extension of this work might be to explore the possibility of using
rotation-invariant split functions in the regression forest to improve its
generalisation capabilities.

\subsection{Timings}
\label{subsec:timings}

\begin{table}[!t]
\centering
\begin{tabular}{ccc}
\toprule
& \textbf{Random Ferns} \cite{Glocker2015,Kaehler2016} & \textbf{Us} \\
\midrule
Per-Frame Training & 0.9ms & 9.8ms \\
Relocalisation     & 10ms  & 141ms \\
\bottomrule
\end{tabular}
\vspace{1mm}
\caption{Comparing the typical timings of our approach vs.\ random ferns during
    both normal operation and relocalisation. Our approach is slower than random
    ferns, but achieves significantly higher relocalisation performance, especially
    from novel poses. All of our experiments are run on a machine with an Intel Core
    i7-4960X CPU and an NVIDIA GeForce Titan Black GPU.}
\label{tbl:timings}
\vspace{-\baselineskip}
\end{table}

To evaluate the usefulness of our approach for on-the-fly relocalisation in new
scenes, we compare it to the keyframe-based random fern relocaliser implemented
in InfiniTAM \cite{Glocker2015,Kaehler2016}. To be practical in a real-time
system, a relocaliser needs to perform in real time during normal operation
(i.e.\ for online training whilst successfully tracking the scene), and ideally
take no more than around $200$ms for relocalisation itself (when the system has
lost track). As a result, relocalisers such as
\cite{Shotton2013,GuzmanRivera2014,Valentin2015RF,Brachmann2016,Massiceti2016},
whilst achieving impressive results, are not practical in this context due to
their need for offline training on the scene of interest.

As shown in Table~\ref{tbl:timings}, the random fern relocaliser is fast both
for online training and relocalisation, taking only $0.9$ms per frame to update
the keyframe database, and $10$ms to relocalise when tracking is lost. However,
speed aside, the range of poses from which it is able to relocalise is quite
limited. By contrast, our approach, whilst taking $9.8$ms for online training
and $141$ms for actual relocalisation, can relocalise from a much broader range
of poses, whilst still running at acceptable speeds. Additionally, it should be
noted that our current research-focused implementation is not heavily optimised,
making it plausible that it could be sped up even further with additional
engineering effort.


\section{Conclusion}

In recent years, offline approaches that use regression to predict
2D-to-3D correspondences
\cite{Shotton2013,GuzmanRivera2014,Valentin2015RF,Brachmann2016,Massiceti2016}
have achieved state-of-the-art camera relocalisation results, but
their adoption for online relocalisation in practical systems such as InfiniTAM
\cite{Kaehler2015,Kaehler2016} has been hindered by the need to train
extensively on the target scene ahead of time.

We show how to circumvent this limitation by adapting offline-trained regression forests to novel scenes online.
Our adapted forests achieve relocalisation performance on 7-Scenes
\cite{Shotton2013} that is competitive with the offline-trained forests of
existing methods, and our approach runs in under $150$ms, making it competitive
in practice with fast keyframe-based approaches such as random ferns
\cite{Glocker2015,Kaehler2016}. Compared to such approaches, we are also much better
able to relocalise from novel poses, freeing the user from manually searching for known
poses when relocalising.

\section*{Acknowledgements}
\noindent
We would like to thank Victor~Prisacariu and Olaf~K{\"a}hler for providing us with the InfiniTAM source code.
\\
This work was supported by the EPSRC, ERC grant ERC-2012-AdG 321162-HELIOS, EPSRC grant Seebibyte EP/M013774/1 and EPSRC/MURI grant EP/N019474/1.

{\small
\bibliographystyle{ieee}
\bibliography{relocalisation}
}

\iftrue
{
	\clearpage
	\appendix
	\twocolumn[\part*{\textsc{Supplementary Materials}} \vspace{2\baselineskip}]

\section{Analysis of Failure Cases}

As shown in the main paper, our approach is able to achieve highly-accurate online relocalisation in under $150$ms, from novel poses and without needing extensive offline training on the target scene.
However, there are inevitably still situations in which it will fail.
In this section, we analyse two interesting failure cases, so as to help the reader understand the underlying reasons in each case.

\subsection{Office}

The first failure case we analyse is from the \emph{Office} scene in the 7-Scenes dataset \cite{Shotton2013}.
This scene captures a typical office that contains a number of desks (see Figure~\ref{fig:officefailure}).
Unfortunately, these desks appear visually quite similar: they are made of the same wood, and have similar monitors and the same associated chairs.
This makes it very difficult for a relocaliser such as ours to distinguish between them: as a result, our approach ends up producing a pose that faces the wrong desk (see Figure~\ref{fig:officefailure}(d)).

On one level, the pose we produce is not entirely unreasonable: indeed, it looks superficially plausible, and is oriented at roughly the right angle with respect to the incorrect desk.
Nevertheless, in absolute terms, the pose is obviously very far from the ground truth.

To pin down what has gone wrong, we visualise the last $16$ surviving camera pose hypotheses for this instance in Figure~\ref{fig:officecandidates}, in descending order (left-to-right, top-to-bottom).
We observe that whilst the top candidate selected by RANSAC relocalises the camera to face the wrong desk, any of the next five candidates would have relocalised the camera successfully.
The problem in this case is that the energies computed for the hypotheses are fairly similar for both the correct and incorrect poses.

Although we do not investigate it here, one potential way of fixing this might be to score the last few surviving hypotheses based on the photometric consistencies between colour raycasts from their respective poses and the colour input image.

\vspace{3cm}\textcolor{white}{.}

\begin{stusubfig}{H}
	\begin{subfigure}{\linewidth}
		\centering
		\includegraphics[width=\linewidth]{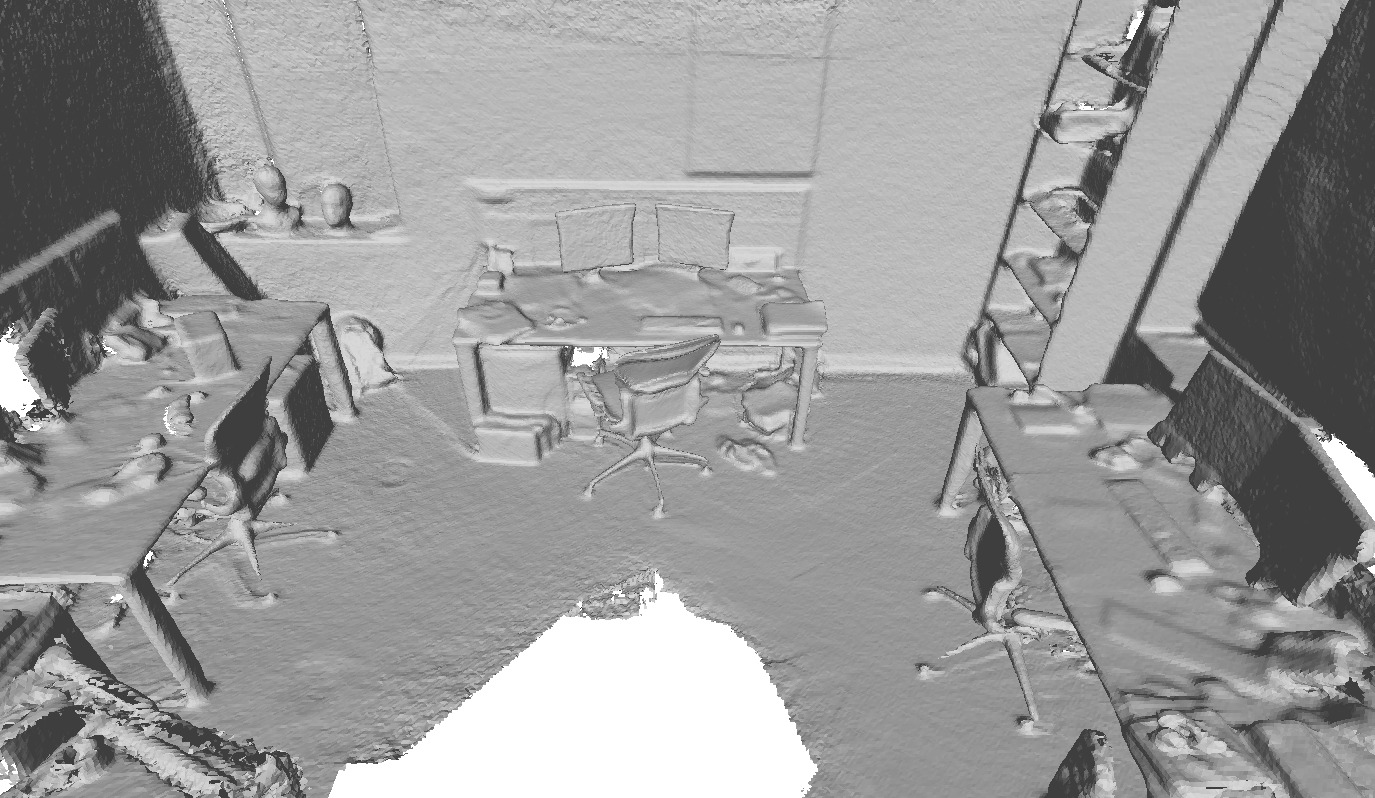}
		\caption{}
	\end{subfigure}%
	\\
	\begin{subfigure}{.49\linewidth}
		\centering
		\includegraphics[width=\linewidth]{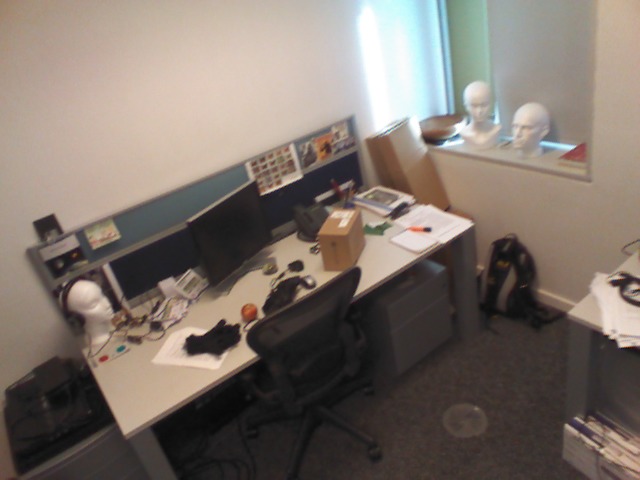}
		\caption{}
	\end{subfigure}%
	\hspace{1.5mm}%
	\begin{subfigure}{.49\linewidth}
		\centering
		\includegraphics[width=\linewidth]{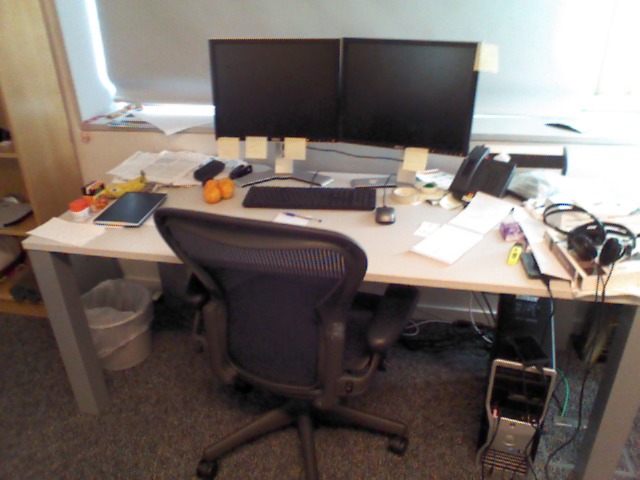}
		\caption{}
	\end{subfigure}%
	\\
	\begin{subfigure}{\linewidth}
		\centering
		\includegraphics[width=\linewidth]{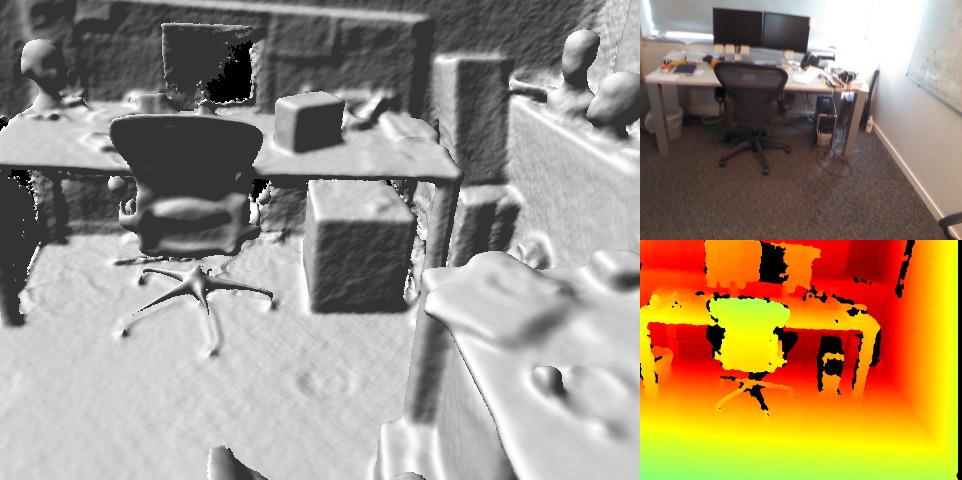}
		\caption{}
	\end{subfigure}%
\caption{The \emph{Office} scene from the 7-Scenes dataset \cite{Shotton2013} (a) contains multiple desks, e.g.\ (b) and (c), that can appear visually quite similar, making it difficult for the relocaliser to distinguish between them. In (d), for example, the relocaliser incorrectly chooses a pose facing the desk in (b), whilst the RGB-D input actually shows the desk in (c).}
\label{fig:officefailure}
\vspace{-1.5\baselineskip}
\end{stusubfig}

\begin{figure*}[!p]
\centering
\includegraphics[height=20.5cm]{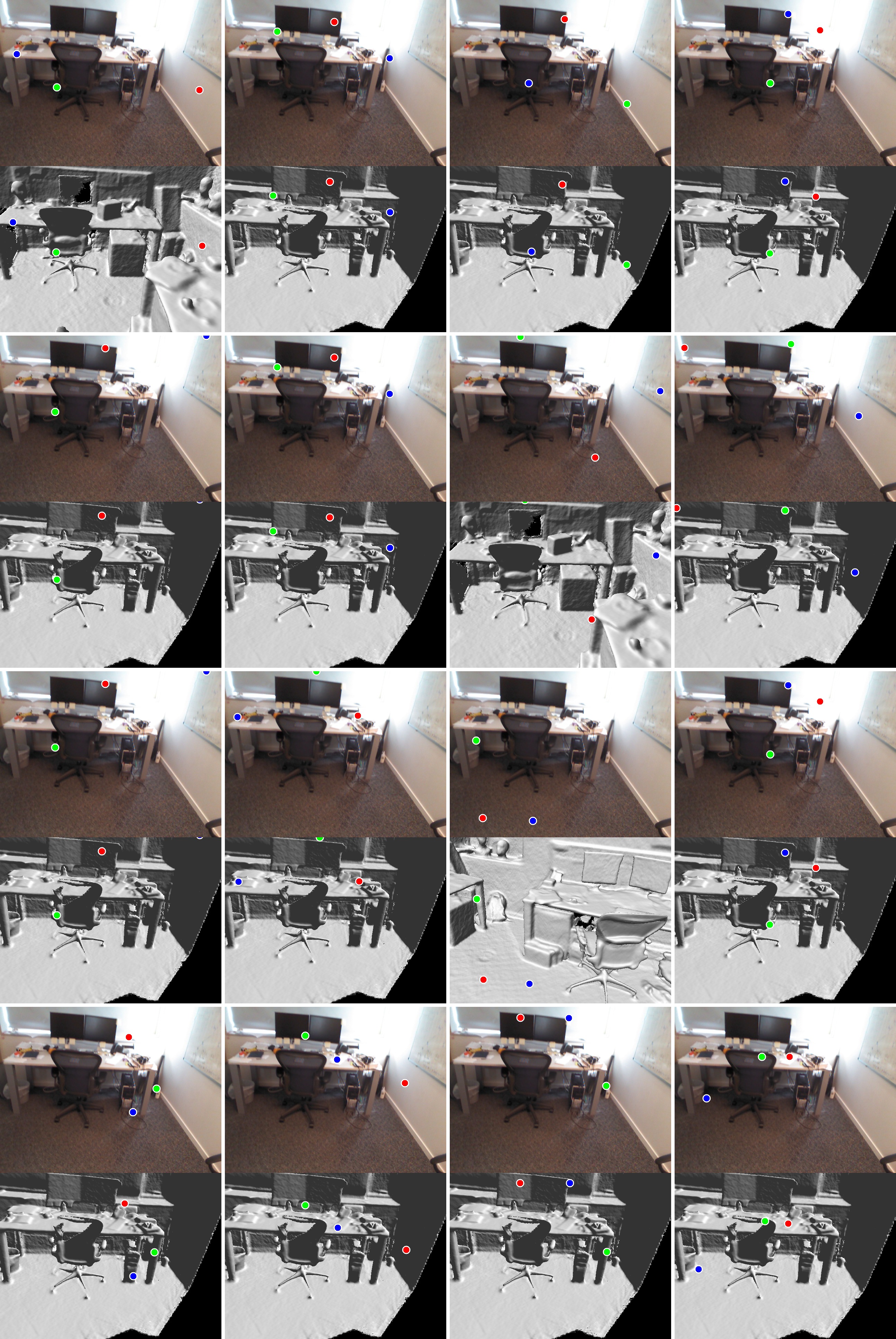}
\caption{The top $16$ pose candidates (left-to-right, top-to-bottom) corresponding to the failure case on the \emph{Office} scene shown in Figure~\ref{fig:officefailure}(d). The coloured points indicate the 2D-to-3D correspondences that are used to generate the initial pose hypotheses. Note that whilst the top candidate selected by RANSAC relocalises the camera to face the wrong desk, any of the next five candidates would have relocalised the camera correctly.}
\label{fig:officecandidates}
\end{figure*}

\clearpage

\subsection{Stairs}

\begin{stusubfig}{!t}
	\begin{subfigure}{\linewidth}
		\centering
		\includegraphics[width=\linewidth]{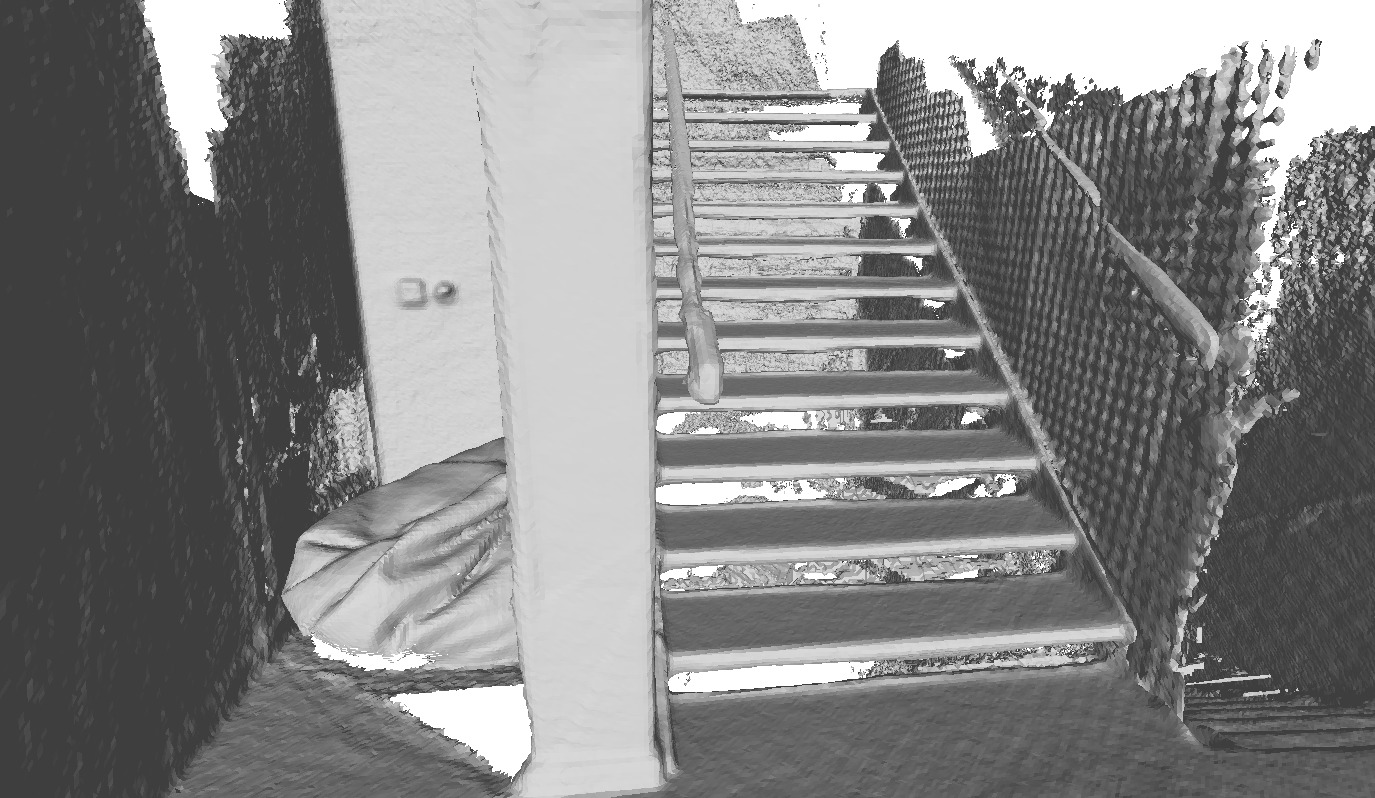}
		\caption{}
	\end{subfigure}%
	\\
	\begin{subfigure}{.49\linewidth}
		\centering
		\includegraphics[width=\linewidth]{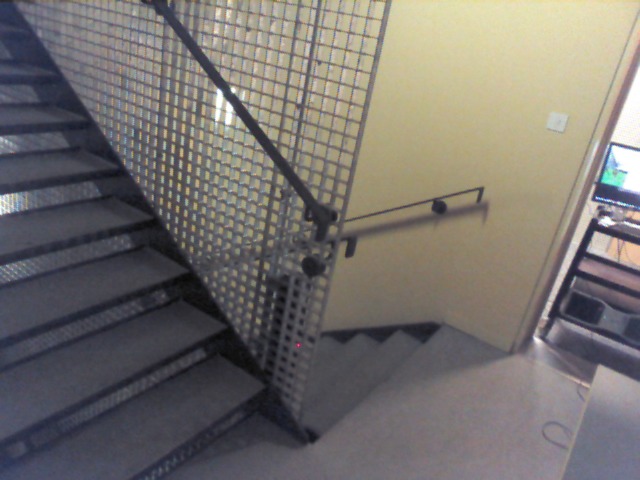}
		\caption{}
	\end{subfigure}%
	\hspace{1.5mm}%
	\begin{subfigure}{.49\linewidth}
		\centering
		\includegraphics[width=\linewidth]{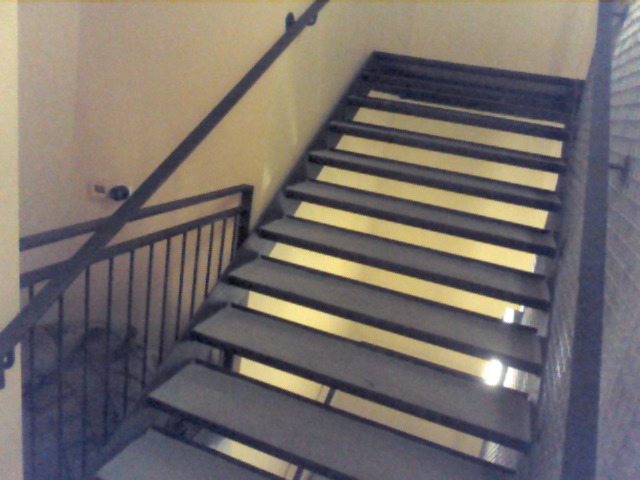}
		\caption{}
	\end{subfigure}%
	\\
	\begin{subfigure}{\linewidth}
		\centering
		\includegraphics[width=\linewidth]{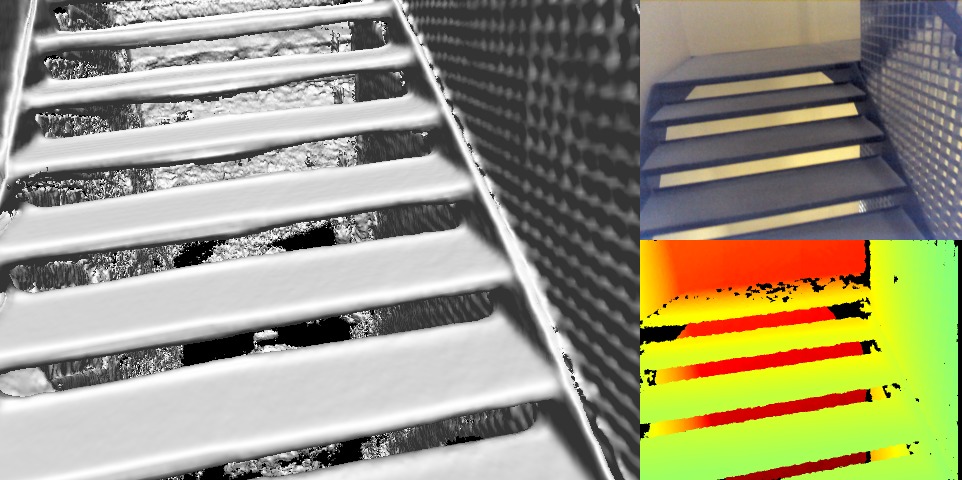}
		\caption{}
	\end{subfigure}%
\caption{The \emph{Stairs} scene from the 7-Scenes dataset \cite{Shotton2013} (a) is notoriously difficult, containing a staircase that consists of numerous visually-identical steps (see (b) and (c)). In (d), many of the 2D-to-3D correspondences predicted by the forest are likely to be of a low quality, since it is hard to distinguish between similar points on different stairs. This significantly reduces the probability of generating good initial hypotheses, leaving RANSAC trying to pick a good hypothesis from an initial set that only contains bad ones.}
\label{fig:stairsfailures}
\vspace{3cm}
\end{stusubfig}

The second failure case we analyse is from the \emph{Stairs} scene in the 7-Scenes dataset \cite{Shotton2013}.
This is a notoriously difficult scene containing a staircase that consists of numerous visually-identical steps (see Figure~\ref{fig:stairsfailures}).
When viewing the scene from certain angles (see Figure~\ref{fig:stairsgood}), the relocaliser is able to rely on points in the scene that can be identified unambiguously to correctly estimate the pose, but from viewpoints such as that in Figure~\ref{fig:stairsfailures}(d), it is forced to use more ambiguous points, e.g.\ those on the stairs themselves or the walls.
When this happens, relocalisation is prone to fail, since the relocaliser finds it difficult to tell the difference between the different steps.

As in the previous section, we can visualise the top $16$ camera pose hypotheses for this instance to pin down what has gone wrong (see Figure~\ref{fig:stairscandidates}).
It is noticeable that in this case, none of the top $16$ hypotheses would have successfully relocalised the camera.
As suggested by the points predicted in the 3D scene for each hypothesis (which are often in roughly the right place but on the wrong stairs), this is because the points at the same places on different stairs tend to end up in similar leaves, making the modes in the leaves less informative (see Figure~\ref{fig:stairsmodes}) and significantly reducing the probability of generating good initial hypotheses.

Unlike in the \emph{Office} case, the problem here cannot be fixed by a late-stage consistency check, since none of the last few surviving hypotheses are of any use.
Instead, one potential way of fixing this might be to improve the way in which the initial set of hypotheses is generated so as to construct a more diverse set and increase the probability of one of the initial poses being in roughly the right place.
An alternative might be to adaptively increase the number of hypotheses generated in difficult conditions.

\clearpage

\begin{stusubfig}{!t}
	\begin{subfigure}{.49\linewidth}
		\centering
		\includegraphics[width=\linewidth]{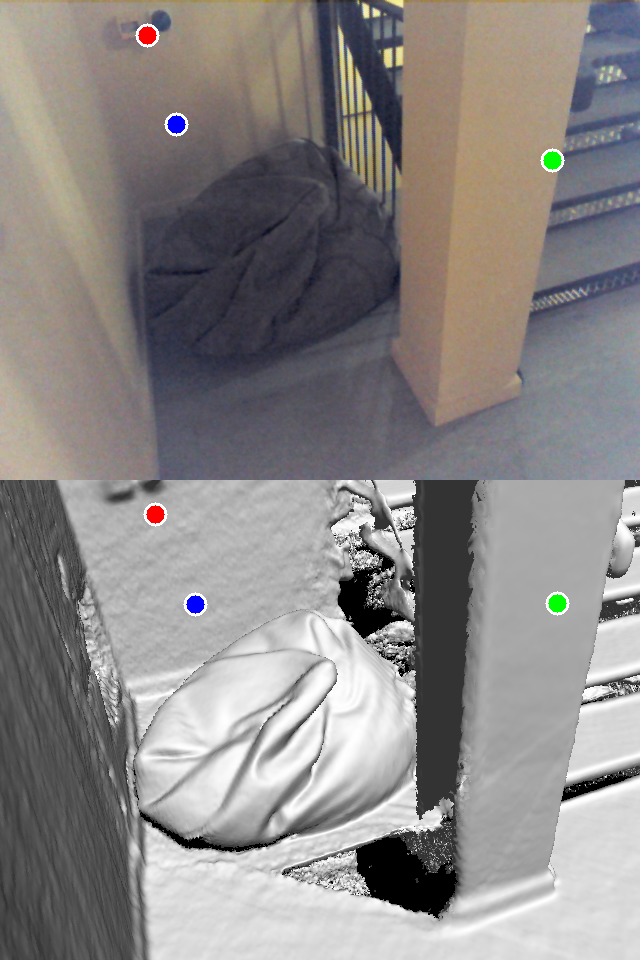}
	\end{subfigure}%
	\hspace{1.5mm}%
	\begin{subfigure}{.48\linewidth}
		\centering
		\includegraphics[width=\linewidth]{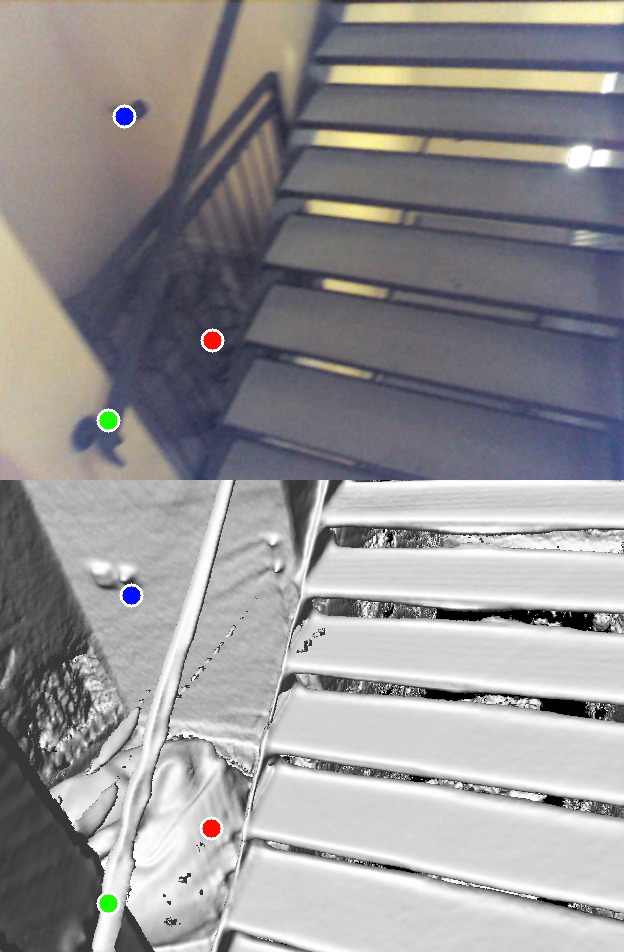}
	\end{subfigure}%
\caption{From certain angles in the \emph{Stairs} scene, the relocaliser is able to rely on points in the scene that can be identified unambiguously to estimate the pose.}
\label{fig:stairsgood}
\vspace{-1.5\baselineskip}
\end{stusubfig}

\begin{figure}[!t]
\centering
\includegraphics[width=\linewidth]{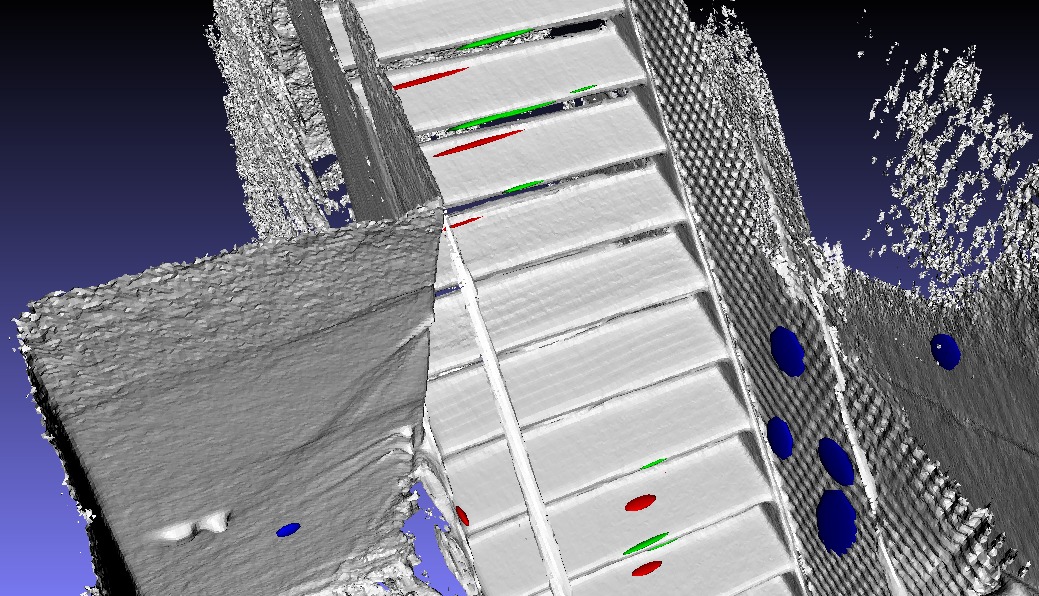}
\caption{The modal clusters contained in the leaves for the points in the optimal camera pose hypothesis from Figure~\ref{fig:stairscandidates}. It is noticeable that points at the same places on different stairs end up in the same leaves, making the distributions in those leaves less informative.}
\label{fig:stairsmodes}
\vspace{6cm}
\end{figure}

\section{Further Successful Examples}

Some further examples of successful relocalisation, this time in the \emph{Fire} scene from the 7-Scenes dataset \cite{Shotton2013}, can be seen in Figure~\ref{fig:firegood}.
As in Figure~\ref{fig:stairsgood}, it is noticeable that the relocaliser tries to rely on points in the scene that can be identified unambiguously where these are available, something that is clearly easier in sequences such as \emph{Fire} that contain many easily-distinguished objects.

\begin{stusubfig}{H}
	\begin{subfigure}{.49\linewidth}
		\centering
		\includegraphics[width=\linewidth]{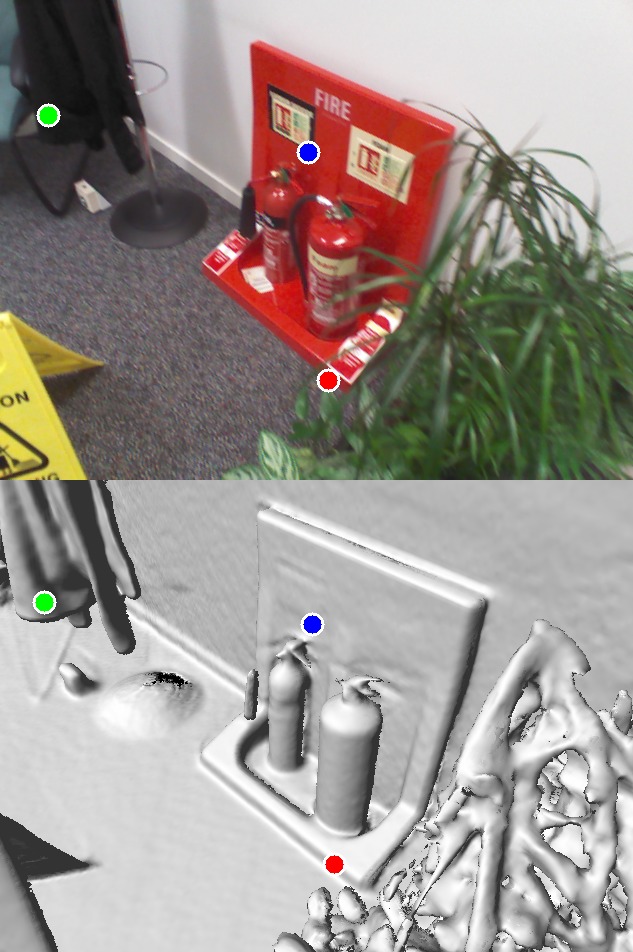}
	\end{subfigure}%
	\hspace{1.5mm}%
	\begin{subfigure}{.49\linewidth}
		\centering
		\includegraphics[width=\linewidth]{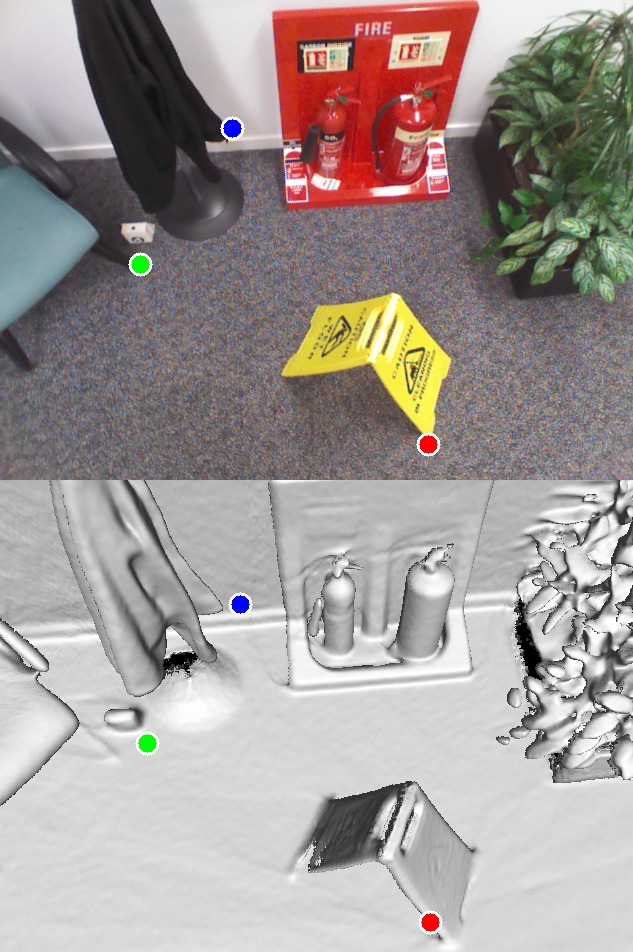}
	\end{subfigure}%
\caption{Further examples of successful relocalisation in the \emph{Fire} scene from the 7-Scenes dataset \cite{Shotton2013}. To estimate the pose, the relocaliser tries to rely on points in the scene that can be identified unambiguously.}
\label{fig:firegood}
\vspace{-1.5\baselineskip}
\end{stusubfig}

\begin{figure*}[!p]
\centering
\includegraphics[height=20.5cm]{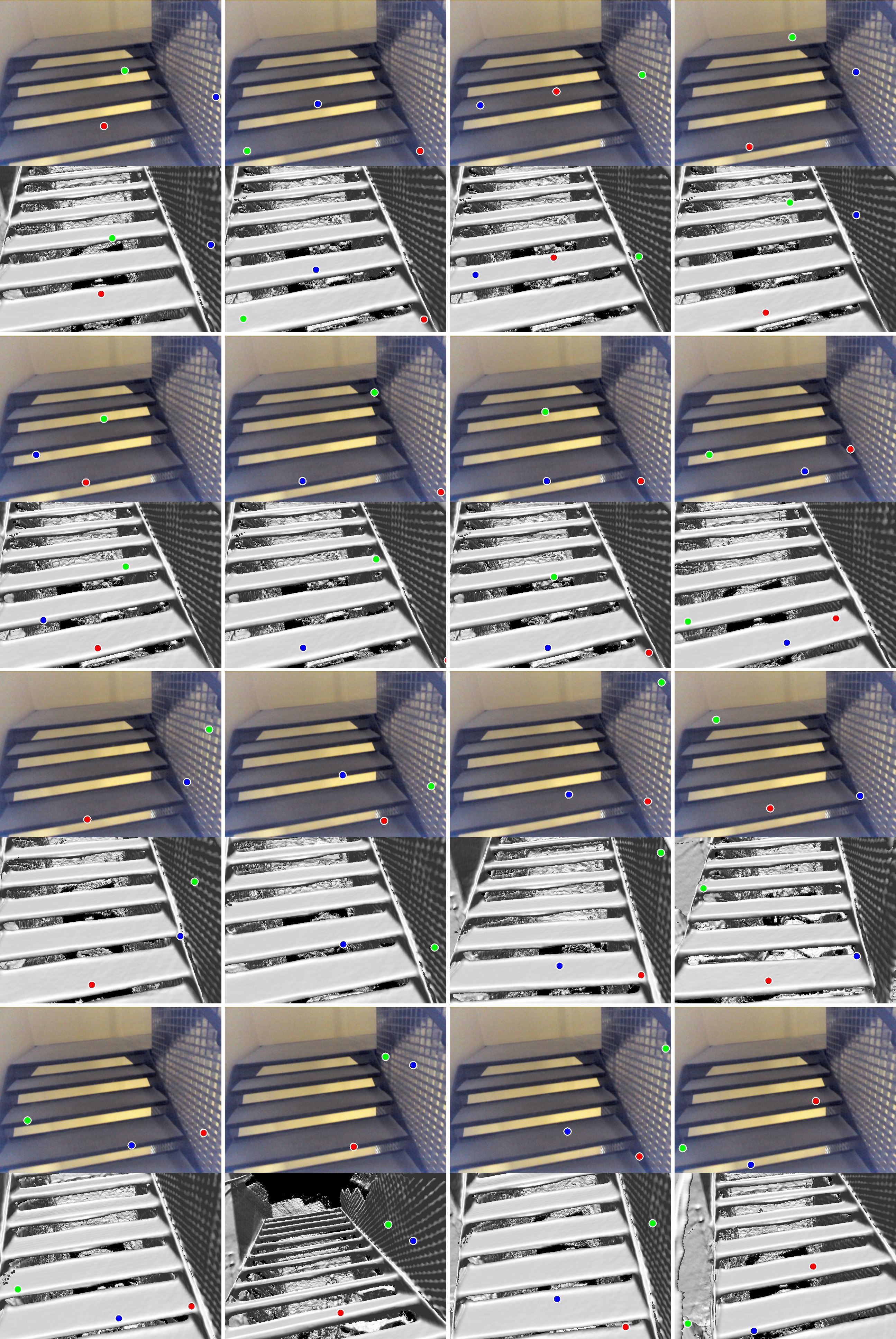}
\caption{The top $16$ pose candidates (left-to-right, top-to-bottom) corresponding to the failure case on the \emph{Stairs} scene shown in Figure~\ref{fig:stairsfailures}(d). The coloured points indicate the 2D-to-3D correspondences that are used to generate the initial pose hypotheses. Note that in this case, none of the candidates would relocalise the camera successfully. This is likely because the points at the same places on different stairs tend to end up in similar leaves, making the modes in the leaves less informative and significantly reducing the probability of generating good initial hypotheses.}
\label{fig:stairscandidates}
\end{figure*}
}
\fi

\end{document}